\documentclass[runningheads]{llncs}
\AddToHook{package/amsmath/before}{%
  \let\vec\relax
  \expandafter\let\csname vec \endcsname\relax
  \DeclareMathAccent{\vec}{\mathord}{letters}{"7E}%
}
\usepackage[T1]{fontenc}
\usepackage{orcidlink}
\AtBeginDocument{%
  \renewcommand{\orcidID}[1]{\unskip\,\orcidlink{#1}}%
}

\usepackage{newtxtext}       %
\usepackage[varvw]{newtxmath}       

\usepackage[all]{nowidow}

\usepackage{graphicx}
\usepackage{xcolor}
\usepackage{algorithm}
\usepackage{algorithmic}
\usepackage{tikz}
\usepackage{adjustbox}
\usepackage{hyphenat}
\usepackage{pifont}
\usepackage{newfloat}
\usepackage{listings}
\usepackage{booktabs}
\usepackage{enumitem}
\usepackage{multirow}
\usepackage{array}
\usepackage{float}
\usepackage{bbding}
\usepackage{diagbox}
\usepackage{forest}

\usetikzlibrary{trees,arrows.meta,positioning}

\makeatletter
\AddToHook{selectfont}{%
  \ifx\f@family\ttdefault
    \hyphenchar\font=\defaulthyphenchar\relax
  \fi
}
\makeatother

\newcommand{\tikzcmark}{\textcolor{green!50!black}{\checkmark}}
\newcommand{\tikzxmark}{\textcolor{red}{\ding{55}}}

\setlist{topsep=2pt, itemsep=1pt, parsep=0pt, partopsep=0pt}
\usepackage{hyperref}
\hypersetup{%
  pdftitle={Formalizing the Moral Evaluation of Speech Acts: Truthfulness, Lies and Ethical Dilemmas},%
  pdfauthor={Benjamin Icard; Jeanne Bonnaventure; Gauvain Bourgne; Jean-Gabriel Ganascia}%
}

\begin{document}
\title{Formalizing the Moral Evaluation of Speech Acts:\texorpdfstring{\\}{ }Truthfulness, Lies and Ethical Dilemmas}
\titlerunning{Formalizing the Moral Evaluation of Speech Acts}
\author{Benjamin~Icard\inst{\text{\Envelope}}\orcidID{0009-0005-4530-5646}
\and
Gauvain~Bourgne\orcidID{0000-0002-5104-9352} \and\\
Jeanne~Bonnaventure\orcidID{0009-0003-7188-9230} \and
Jean-Gabriel~Ganascia\orcidID{0000-0002-8181-1379}}
\authorrunning{Icard et al.}
%
\institute{LIP6, Sorbonne University, CNRS, France \\
\Envelope\ Corresponding author: \email{benjamin.icard@lip6.fr}}
\maketitle              
\begin{abstract}
In life-or-death situations, a benevolent lie may appear more moral than telling the truth. Yet such lies can backfire, producing unintended and sometimes fatal consequences. This tension, famously disputed by Kant and Constant in 1797, applies not only to lying but to assertive speech acts in general, raising the question of which utterance should be chosen when moral stakes are high. We present a logical framework for the ethical evaluation of speech-act utterances based on agents' beliefs. Implemented in Answer Set Programming (ASP), the framework assesses utterances under deontologism, consequentialism, and principialism, and is illustrated on Sartre's \textit{The Wall} (1939), a reworking of that controversy in which lying leads alternately to rescue and to death. Our setting is general by design: as two variants show, accommodating a new moral situation amounts to adjusting parameters, not rules.

\end{abstract}


\section{Introduction}

Moral theories aim to explain what makes actions right or wrong and to guide agents towards correct moral verdicts \cite{timmons2022moral}. In machine ethics, these approaches and their formalization emphasized \textit{physical actions}, through dilemmas like the Trolley Problem \cite{foot1967abortion} or the Transplant case \cite{thomson1985trolley}. By contrast, \textit{verbal actions}, more precisely \textit{speech act utterances} such as lying, truth-telling, or withholding information, have received far less systematic attention \cite{fogal2018new,cuneo2020}. Yet in moral life as in multi-agent systems, agents act on one another through their utterances, which can be honest or deceptive, and beneficial or harmful.

The moral status of such utterances, and of lying above all, is a classical concern of moral philosophy \cite{bok1978lying}, famously disputed by Constant~\cite{constant1988constant} in 1796, for whom
honesty is owed only to those who deserve the truth, and by
Kant~\cite{kant1996} in his reply of 1797, for whom honesty is an
unconditional duty. This controversy widens into the divide between
\textit{internalist} theories, judging an action by the agent's intentions, as in Kant's deontologism, and \textit{externalist} ones, judging it by its effects, as in the consequentialism of Bentham and Mill
\cite{Bentham1789,Mill1863}. Constant's principialism stands as a
compromise between the two, weighing intentions while admitting exceptions. Beyond the permissibility of lying, this debate asks which speech acts, truthful assertion, lie, or silence, are ethical, towards whom, and on what grounds.

This paper proposes a logic-based framework to conduct such an
ethical evaluation of actions, speech-act utterances included, under the
classical moral theories. Speech-act theory is epistemic by design,
emphasizing the speaker's honesty and the truth value of the utterance
\cite{searlevanderveken1985,fipa2002}, but, as actions, utterances also
trigger outcomes, possibly indirect, that can be beneficial or detrimental.
For their proper evaluation, we introduce (i) a simplified agent architecture, based on beliefs and equipped with an explicit
decision theory; (ii) a causal model of actions, whether physical or verbal,
deriving their outcomes; (iii) an epistemic layer capturing assertive speech
acts and their types; and (iv) a unified formalism to represent deontologism
and consequentialism, with principialism as a mixed approach. Our contribution is a reusable architecture for comparing these moral theories, not an exhaustive formalization of each.

We realise the framework in Answer Set Programming
(ASP)~\cite{gelfondlifschitz1991classical,lif08}, well suited to default norms and their exceptions.

We illustrate the debate on the ethics of
speech acts using Sartre's \textit{The Wall}
(1939)~\cite{sartre1939mur}: interrogated by Falangists during the Spanish Civil War, the anarchist Pablo Ibbieta lies twice about where his comrade Ramon Gris is hiding. The first lie is beneficial and protects Ramon, but the second, being true by tragic coincidence, is detrimental and leads to his arrest and death. The story has been read as a literary treatment of the Kant-Constant controversy, the second lie instantiating Kant's responsibility argument~\cite{sweeney1985lying}. The implementation derives the epistemic type of each lie and returns each moral theory's verdict. We also use variants to show the generality of the framework and its sensitivity: minimal parameter changes make converging theories diverge.

Section~\ref{sec:extant} reviews prior work and shows that speech acts still lack a systematic ethical evaluation. Section~\ref{sec:concepts} sets out the conceptual framework, the epistemic typology of utterances and the three moral theories with their variants, applied to \textit{The Wall}. Section~\ref{sec:causalmod} develops the formal framework, from its typed vocabulary and the \texttt{utter} predicate to its evaluators. Section~\ref{sec:evalframework} reports the ASP evaluation. Section~\ref{sec:discussion} discusses two alternatives of Sartre's \textit{The Wall}, and Section~\ref{sec:conclusion} concludes and sets the basis for future work.


\section{Motivations}
\label{sec:extant}
\subsection{Moral Speech Acts}
\label{ssec:moralspeech}

Speech acts are a distinctive class of actions, performed by uttering rather than by acting physically on the world~\cite{austin}. Searle's taxonomy of five illocutionary classes~\cite{searle1969speech,searle1976} was cast in illocutionary logic by Searle and Vanderveken~\cite{searlevanderveken1985} and, in multi-agent systems, operationalised by agent communication languages, most prominently FIPA-ACL~\cite{fipa2002,labrou1999}, the standardised successor of KQML~\cite{finin1994}. Their moral evaluation, however, is left aside: these languages fix a library of performatives, among which the two communicative acts central to lying are the assertive \texttt{inform} and the directive \texttt{request}, with questions being a special case of the latter.

The moral evaluation of lying is exemplified by the \textit{lying controversy} opposing Constant and Kant. In \cite{constant1988constant}, Constant held that an unconditional duty of truthfulness would be absurd, since truth is owed only to those who have a right to it, which a murderer asking where his victim is hidden forfeits, a form of moral contextualism~\cite{rousseliere2018political} that we
call ``principialism''. Replying in \cite{kant1996}, Kant maintained that truthfulness admits no exception, even towards a murderer, the hallmark of ``deontologism''~\cite{alexander2016deontological}, and further argued that the liar is answerable for the resulting harm, since the speaker controls the sincerity of his utterance but not its truth. Notably, Constant's principialism differs from consequentialism~\cite{sinnott2015consequentialism}, which judges strictly by outcomes. Yet moral theories have been developed mainly for physical actions, as in the Trolley Problem~\cite{foot1967abortion,thomson1985trolley}, and far less for verbal ones, so the controversy was never given a systematic formal treatment.

The objects of lies, and of speech acts more generally, are \textit{utterances}. Epistemically, an utterance is \textit{honest} or \textit{dishonest} according to whether the speaker believes its content, and \textit{true} or \textit{false} according to the facts~\cite{Grice1957,grice1975logic}, a lie being a dishonest utterance meant to deceive~\cite{mahon2008definition}. Crossing the two dimensions yields epistemic typologies that distinguish \textit{objective lies}, dishonest and false~\cite{coleman1981prototype}, from \textit{erroneous lies}, dishonest yet true~\cite{wiegmann2016lying}. Such typologies, however, have rarely been given a fine-grained ethical evaluation \cite{bok1978lying,carson2010lying}.

\subsection{Logical Models of Moral Actions}

Logic-based approaches to moral reasoning have formalised through preference~\cite{hansson2001structure}, epistemic~\cite{lorini2015logic}, and defeasible~\cite{dennis2021defeasible} logics, as well as BDI models of deception~\cite{ganascia2015}, higher-order logic~\cite{benzmuller2020designing}, and planning~\cite{lindner2019moral}. Among implementations, Answer Set Programming (ASP)~\cite{gelfondlifschitz1991classical} is well suited to non-monotonic reasoning and conflicting norms, and has modelled physical actions via event calculus for causality and responsibility~\cite{berreby2015modelling}, modular ethical principles~\cite{berreby2017declarative}, and concurrency and omissions in the ACE framework~\cite{bourgne2021ace}. Verbal actions, whose evaluation also turns on what the speaker believes, call instead for a model of communication.

Such a model is provided by FIPA-ACL, whose semantics is mentalistic: the feasibility precondition of an assertive such as \texttt{inform} requires the speaker to believe the asserted content~\cite{pittmamdani1999}, so a well-formed assertion is sincere, hence truthful, by construction. As this constrains private mental states rather than observable behaviour, it is not externally verifiable~\cite{singh1998}, which has prompted variants recasting the semantics in terms of social commitments~\cite{fornaracolombetti2002} or public grounding~\cite{gaudou2006}. Modelling lying then requires relaxing the sincerity precondition to represent the divergence between what a speaker believes and what it asserts, as in dynamic epistemic logics of lying~\cite{vanditmarsch2014dynamics}. Logical accounts of dishonest communication~\cite{sakama2010logical,sakama2015formal} address such epistemic aspects without integrating ethical theories.

\subsection{Sartre's Critical Example}

Sartre's \textit{The Wall}~\cite{sartre1939mur} illustrates the difficulty. The
anarchist Pablo Ibbieta, interrogated by Falangists during the Spanish Civil
War, must decide whether to reveal where his comrade Ramon Gris is hiding. He believes Ramon at his family's house, whereas Ramon is in fact at the cemetery. He lies twice. Naming Madrid, he asserts what he believes false and what is false, an \textit{objective lie} that protects Ramon. Pressed again, he names
the cemetery, deceptive once more but true this time, an \textit{erroneous lie} leading to Ramon's capture and death.\footnote{The two-round
Madrid--cemetery questioning used throughout is a stylized reconstruction of Sartre's narrative, adopted for illustration.} One and the same dishonest but
benevolent intention thus yields opposite outcomes. The story transposes the Kant-Constant controversy of Section~\ref{ssec:moralspeech}: Pablo faces
Constant's murderer demanding the whereabouts of his victim, and the second lie
realizes Kant's responsibility argument, proving true and harming the person it
was meant to protect. Sweeney indeed argues that Sartre built the story on the examples of Kant's reply to Constant~\cite{sweeney1985lying}.

These strands remain largely disjoint. Typologies of lying seldom receive fine-grained ethical evaluation, logical models of moral action target physical rather than verbal actions, accounts of dishonest communication set ethics aside, and cases such as Sartre's lack systematic formal treatment. We introduce a formal framework to help close these gaps.


\section{Conceptual Setting}
\label{sec:concepts}

This section first characterizes utterances epistemically, by the speaker's honesty and the truth of their content, then presents the three moral theories and their variants, which further weigh the speaker's motive and the resulting outcome. Sartre's \textit{The Wall} finally serves to compare their verdicts, as one instance of the Kant-Constant problem: what a speaker owes, and to whom.

\subsection{Epistemic Speech Acts}
\label{ssec:epistespeech}

Epistemically, speech-act utterances are characterized along two dimensions: the speaker's intention, \textit{honest} when speakers assert what they believe to be true and \textit{dishonest} when they assert what they do not, and their
semantic value, \textit{true} when they match objective facts and \textit{false} when they contradict them. Epistemic externalist theories prioritize truth status while internalist ones emphasize speaker intention, in line with Grice's
theory of meaning and communication \cite{Grice1957,grice1975logic}.
Table~\ref{tab:speech} crosses the two dimensions in a 2$\times$2 matrix, each
combination aligned with established linguistic labels
(see e.g.,~\cite{icardthesis2019}). Although alternative terminologies exist (see e.g.,~\cite{Fallis2014}), we retain these four. The formal model uses belief-content agreement as a proxy for epistemic intention, without representing an intention to deceive \textit{per se}.

\begin{table}
\caption{Epistemic types of speech acts based on the speaker's
communicative intention and the content of their utterances.}
\label{tab:speech}
\centering
\small
\begin{tabular}{@{}lc!{\hspace{1.5em}}c@{}}
\toprule
\multirow{2}{*}{\textbf{Speaker's Intention}} & \multicolumn{2}{c}{\textbf{Content of the Utterance}} \\
\cmidrule(l){2-3}
 & True & False \\
\midrule
Honest    & \textit{objective truth} & \textit{erroneous truth} \\
Dishonest & \textit{erroneous lie}   & \textit{objective lie} \\
\bottomrule
\end{tabular}
\end{table}

\subsection{Moral Theories of Speech Acts}
\label{ssec:moraltheo}

From a moral standpoint, both the speaker's epistemic intention and the truthfulness of the utterance matter, but two further dimensions also bear on evaluation. The first is the speaker's \textit{motive}, \textit{benevolent} when it aims to protect, as in saving a life, or \textit{maleficent} when it aims to harm, as in enabling a death. The second is the resulting \textit{outcome}, \textit{beneficial} when its consequences are positive for those concerned or \textit{detrimental} when they are negative.

Internalist theories weigh mainly the speaker's intention and motives, externalist ones the outcomes of the utterance, with subtleties within each. Within internalism, \textbf{deontologism}, in particular Kant's, evaluates the morality of the speaker's moral motive and epistemic intention, considering two maxims as mandatory \cite{kant1998,kant1997}:

\begin{itemize} 
\item[—] \texttt{maxim1}: the speaker's motive is connected to the absolute duty to avoid \textit{maleficence}, which implies refraining from enabling murder.
\item[—] \texttt{maxim2}: the speaker's intention is linked to the unconditional duty to be \textit{honest}, which entails always stating what they believe to be true.

\end{itemize}

Concerning \textbf{principialism}, Constant's principialism \cite{constant1988constant} requires the speaker to have honest epistemic intentions and no maleficent motives. However, unlike deontologism, these duties are not absolute maxims but principles that may admit exceptions. We identify two such principles, each giving rise to a distinct variant of principialist reasoning:

\begin{itemize}
\item[—] \texttt{principle1}: Honesty is a default duty, but dishonesty is allowed if the interlocutor does not deserve the truth, as in cases of murderous intent or maleficence. This defines \texttt{\small principialism1}.

\item[—] \texttt{principle2}: Honesty remains the default norm, but dishonesty is permitted if, under counterfactual reasoning, telling the truth would result in worse consequences. This defines \texttt{\small principialism2}.
\end{itemize}

Within \textbf{consequentialism}, we can also distinguish between two approaches aligned with the utilitarian theory proposed by \cite{Bentham1789} and expanded by \cite{Mill1863}: expected utilitarianism, labeled here as \texttt{\small consequentia\-lism1}, and real utilitarianism, labeled as \texttt{\small consequentia\-lism2}. On the one hand, \texttt{\small consequential\-ism1} evaluates the morality of speech-act utterances, based on their anticipated outcomes at the time of decision-making, emphasizing the speaker’s motives as being either benevolent or maleficent. On the other hand, \texttt{\small consequent\-ialism2} assesses speaker's utterances according to their actual outcomes, judging whether they are ultimately beneficial or detrimental, regardless of the agent’s anticipations.

\begin{table}[ht]
\centering
\caption{Moral evaluation of Pablo's speech-act options in Sartre's
\textit{The Wall}, where \textbf{m}, \textbf{p}, \textbf{c}
abbreviate \texttt{maxim}, \texttt{principialism},
\texttt{consequentialism}; with \tikzcmark: permissible, and \tikzxmark:
impermissible. \textbf{NB}: here, verdicts assume no better alternative (silence, evasion) is
available to Pablo.}

\label{tab:eval}
\setlength{\tabcolsep}{1.6pt}
\scriptsize

\resizebox{\textwidth}{!}{%
\begin{tabular}{@{}lccc*{7}{c}@{}}
\toprule
\multirow{2}{*}{\textbf{Utterance}} & \multicolumn{2}{c}{\textbf{Pablo's Plan}}
& \multicolumn{1}{c}{\textbf{Outcome}}
& \multicolumn{3}{c}{\textbf{Deontologism}}
& \multicolumn{2}{c}{\textbf{Principialism}}
& \multicolumn{2}{c}{\textbf{Consequentialism}} \\
\cmidrule(lr){2-3} \cmidrule(lr){4-4} \cmidrule(lr){5-7}
\cmidrule(lr){8-9} \cmidrule(l){10-11}
& Intention & Motive
& For Ramon
& \textbf{m1} & \textbf{m2} & \textbf{m1+2}
& \textbf{p1}  & \textbf{p2} 
& \textbf{c1}  & \textbf{c2} \\

& \multicolumn{3}{c}{\textit{Type of Evaluation}} & \multicolumn{3}{c}{\textit{Absolute}} & \textit{Local}  & \textit{Global}
& \textit{Anticipations} & \textit{Reality} \\

\midrule
\texttt{objective\_truth}
& Honest & Maleficent & Detrimental
& \tikzxmark & \tikzcmark & \tikzxmark
& \tikzxmark & \tikzxmark
& \tikzxmark & \tikzxmark \\
\texttt{erroneous\_truth}
& Honest & Maleficent & Beneficial
& \tikzxmark & \tikzcmark & \tikzxmark
& \tikzxmark & \tikzxmark
& \tikzxmark & \tikzcmark \\
\texttt{objective\_lie}
& Dishonest & Benevolent & Beneficial
& \tikzcmark & \tikzxmark & \tikzxmark
& \tikzcmark & \tikzcmark
& \tikzcmark & \tikzcmark \\
\texttt{erroneous\_lie}
& Dishonest & Benevolent & Detrimental
& \tikzcmark & \tikzxmark & \tikzxmark
& \tikzcmark & \tikzcmark
& \tikzcmark & \tikzxmark \\
\bottomrule
\end{tabular}}
\end{table}

Table~\ref{tab:eval} applies these criteria to Pablo's four options: with both maxims in force, deontologism prohibits all of them, principialism permits both lies (\texttt{\small erroneous\_lie}, \texttt{\small objective\_lie}), by murderous intent (\texttt{\small principialism1}) or by
counterfactual comparison (\texttt{\small principialism2}), as does \texttt{\small consequentialism1}, whereas \texttt{\small consequentialism2} permits only one of the two lies (\texttt{\small objective\_lie}).

The theories thus differ not only in
their permissibility criteria, but in which distinction, intention, motive or outcome, they treat as decisive. Two conventions apply: no better alternative, such as silence or a later evasion, is assumed available, and each utterance is judged independently, whereas \texttt{\small consequentialism2} aggregates a whole decision and \texttt{\small principialism2} compares complete situations, so verdicts may change once utterances combine with other actions, as in the ASP evaluation of
Section~\ref{sec:evalframework}.

\section{Logic of Speech Acts Ethics}
\label{sec:causalmod}

In this section, we formalize the logical language and architecture of our framework, designed to generate and evaluate ethically decisions that can involve speech acts utterances.

\subsection{Overview}

Our language is first-order in syntax but propositional in expressive power: we consider finitely many grounded terms, so the Herbrand base is finite and we treat grounded atomic formulas as propositional variables.

As the framework must allow (i) derivation of the consequences of the agent's decision and (ii) evaluation of the sincerity and veracity of its utterances, the architecture is built on a causal theory, whose input are determined by a (possibly incorrect) belief base and a decision theory. This causal theory, inspired by Structural Equation Models \cite{pearl2009causality} and recast as logic-programming rules, is structured by variable dependencies, alongside the distinctions \textit{endogenous}/\textit{exogenous},
\textit{controlled}/\textit{uncontrolled}, and \textit{certain}/\textit{uncertain}. It derives, deterministically given the context and the agent's beliefs, the consequences of the choices, speech-act utterances included, of the sole agent under moral evaluation, the \emph{decider} \texttt{d}. This setting is deliberately minimal: as we evaluate a single decider, sincerity and veracity reduce to checking an asserted content against a belief base (the agent's or the environment's), so no nesting of belief modalities is needed.

\subsection{Basic Language}
\label{ssec:basiclanguage}

\subsubsection{Vocabulary.}
We define four disjoint sets of fixed-arity atoms $\mathcal{I}$, $\mathcal{C}$, $\mathcal{A}$, $\mathcal{E}$. 

$\mathcal{I}$ (\textit{invariable predicates}) encode structural properties of the domain, such as domain predicates (e.g., \texttt{Person}, \texttt{Location}) and comparison predicates (e.g., $\leq$, $=$). Their truth is fixed across all states and treated as common knowledge. These invariants are typically compiled away at instantiation time.

\(\mathcal{C}\) (\textit{context variables}) represent uncertain knowledge about the state of the system not controlled or influenced by the agent's decision. We call the interpretation of these variables a \textit{belief base}. They represent the main input (exogenous variables) of the causal model.

\(\mathcal{A}\) (\textit{action predicates}) encode the agents' deliberate actions and reactions. They define the choice space and are the object of ethical assessment: their elements \emph{enter into the agent's decision}, being controlled by the agent, while their feasibility and outcomes may be context-constrained. We call the interpretation of these variables a \emph{decision}.
They are considered input (exogenous variable) of the causal theory, but possible decisions can in practice be constrained by the other kinds of variables through the decision theory.
They include communicative acts issued by the agent whose decision we are evaluating, i.e. the decider.

\(\mathcal{E}\) (\textit{event predicates}) represent the consequence unfolding from the agent's choices in the given context. 
They are endogenous variables, uniquely determined by a belief base and a decision through the causal theory. 
As such, they are uncertain through $\mathcal{C}$, and only indirectly controllable through \(\mathcal{A}\). Unlike actions, events \emph{do not enter the agent's decision}: they occur automatically or are fixed externally. They model all events the environment triggers in reaction to the decision. Since only the decider's choices are assessed, actions of other agents are treated as part of this reaction and included in $\mathcal{E}$.

These atomic events represent belief- or context-relative transitions. Complex epistemic and moral classifications of speech acts are defined as composite predicates in subsequent sections.

\subsubsection{The Utterance Predicate.}

The communicative acts within \(\mathcal{A}\) and \(\mathcal{E}\) share a single form. Drawing from \cite{sakama2010logical,sakama2017causal}, we use the umbrella predicate \texttt{utter} to represent them, packaging some message meta-information with the message itself, given as a performative of FIPA-ACL~\cite{fipa2002} applied to a logical content.

\begin{center}
\texttt{utter(X,Y,}\textit{N}\texttt{,$\varphi(\sigma)$)}
\end{center}
Its components are the \emph{speaker} $X$, the \emph{addressee} $Y$, the \emph{round} \textit{N} (a set of conversation identifiers which allows one to link questions to answers), the \emph{illocutionary force} $\varphi$ (a performative), and the \emph{content} $\sigma$. 
Thus, given decider \texttt{d}, \texttt{utter(d,X,$\cdot$,$\cdot$)} and \texttt{utter(X,d,$\cdot$,$\cdot$)} encode the two conversational directions, first one being treated as an action (in $\mathcal{A}$) and second one as an event (in $\mathcal{E}$). 
The schema follows classical illocutionary logic, in which an act is a force applied to a propositional content~\cite{searlevanderveken1985}, and additionally records the message meta-information alongside $\varphi$ and $\sigma$.

In our example, two communicative acts instantiate this schema, differing only in force and in the type of content:
\begin{itemize}
\item[—] \textit{Informing} (in \(\mathcal{A}\)): \texttt{utter(X,Y,}\textit{N}\texttt{,inform($\sigma$))} by which $X$ asserts a $\sigma$ content to $Y$, with $\sigma\in\mathcal{C}\cup\mathcal{E}$; 

\item[—] \textit{Question} (in \(\mathcal{E}\)): \texttt{utter(X,Y,}\textit{N}\texttt{,query($\sigma(x)$))} by which $X$ asserts a $\sigma(x)$ question to $Y$, with $x$ a free variable that an answer should bind to a value.

\end{itemize}

\subsubsection{Instantiation in Sartre's \textit{The Wall}.}
Sartre's \textit{The Wall} runs throughout as our example. It involves three persons, now denoted by constants: \texttt{p} for Pablo (the prisoner deciding what to say), \texttt{q} for the Falangists (his interrogators), and \texttt{r} for Ramon (the comrade in hiding). Here the decider (agent whose decisions we are evaluating) is Pablo (\texttt{p}).
The relevant locations are the constants \texttt{family}, \texttt{cemetery}, and \texttt{madrid}. Predicates are stated generically, with variables $X, Y, Z$ for agents, $L$ for locations, $E$ for reasons, $N$ for conversation IDs, and $\sigma$ for utterance contents. Instantiating the four sets and the utterance schema on this scenario yields the following.

\begin{itemize}
\item[$\bullet$] $\mathcal{I}$ \textnormal{(\textit{invariable predicates}):} \texttt{person(X)}, \texttt{location(L)}.

\medskip
\item[$\bullet$] $\mathcal{C}$ \textnormal{(\textit{context variables}):} \texttt{at(X,L)}(indicates the location of a person), \texttt{hostile(Y,Z)} (\texttt{Y} is hostile towards \texttt{Z}), and \texttt{murderer(Y)} (consider \texttt{Y} as being a murderer by nature). Moreover, we also represent as context the predicates \texttt{credible($\sigma$,Y)} which indicates whether \texttt{Y} could find the statement $\sigma$ to be credible if it is presented to him. 
In practice, at instantiation, \texttt{X}=\texttt{r}, \texttt{Y}=\texttt{q}, $\sigma$=\texttt{at(r,L)}, \texttt{Z}$\in\{$\texttt{r},\texttt{p}$\}$.

\medskip
\item[$\bullet$] $\mathcal{A}$ \textnormal{(\textit{action predicates}):} Pablo's initial options are to answer the Falangists' question with an assertive communicative act (of the form \texttt{utter(p,q,}\allowbreak\textit{N}\texttt{,inform(at(r,L)))}), or to keep \texttt{silence(p,q,}\textit{N}). Conditionally, Pablo might also attempt to escape from the Falangists: \texttt{evade(p,q)}.

\medskip
\item[$\bullet$] $\mathcal{E}$ \textnormal{(\textit{event predicates}):} the Falangists' interrogation is the reference question \texttt{utter(q,p,}\textit{N}\texttt{,query(at(r,$x$)))}, whose answer binds $x$. Depending on Pablo's reply, non-communicative events may then occur: 

\smallskip
\begin{itemize}
\item[—] \texttt{check(X,Y,L,Z)} (\texttt{X} checks for \texttt{Z} at location \texttt{L} based on \texttt{Y}'s information),
\item[—] \texttt{kill(X,Y,Z)} (\texttt{X} kills \texttt{Z} based on \texttt{Y}'s information),
\item[—] \texttt{harm(X,Y,E)} (\texttt{X} harms \texttt{Y} for reason \small{\texttt{E}$\in$\{\texttt{uncredibleStatement},

\texttt{refusalToAnswer},\texttt{uncoveredLie}, 
\texttt{failedEscape}\}}), 
\item[—] \texttt{escaped(X,Y)} (\texttt{X} successfully escaped from \texttt{Y}).
\end{itemize}
\end{itemize}

\subsection{Logical Architecture}
\label{ssec:logarch}

Our framework aims at providing an ethical evaluation of the choices of an agent with uncertain knowledge in a situation involving assertive speech acts with practical consequences. 
The first layer, described in this section, is concerned with determining the possible choices and determining both their expected and real consequences. 
A second evaluative layer, described next, assesses the sincerity and veracity of each assertion and is used to evaluate first locally, then globally the different decisions according to different ethical principles.

At the basis of this first layer are \textit{Belief Bases}.
They are built upon context predicates ($\mathcal{C}$) and thus represent either the individual knowledge of the decider or the truth state of the environment. We write $\mathcal{B}_{\texttt{env}}$ for the environment's belief base and $\mathcal{B}_{a}$ for that of an agent $a$. In both cases $\mathcal{B}\in 2^{\mathcal{C}}$.

\textit{Causal Theory}, written $\Delta$, is the set of causal rules treated as certain knowledge about the situation's dynamic. Causal rules have the form:
\[
\varepsilon \leftarrow \delta
\]
where the head $\varepsilon$ depends on both $\mathcal{B}_{a}$ and $D$, and the body $\delta$ is a conjunction of literals,
\[
\delta = \theta \wedge \kappa \wedge \alpha \wedge \varepsilon'
\]
with $\varepsilon \in \mathcal{E}$ and $\theta$, $\kappa$, $\alpha$, $\varepsilon'$ are conjunctions of literals form respectively $\mathcal{I}$, $\mathcal{C}$, $\mathcal{A}$ and $\mathcal{E}$. $\theta$ is set to $\top$ or $\bot$ and compiled away at instantiation), while the interpretation of $\kappa$ and $\alpha$ is given by a belief base $\mathcal{B}$ and a decision $D$. We adopt a closed-world interpretation: atoms absent from the relevant belief base, decision, or derived event set are false. In our rule bodies, $\neg$ tests this falsity and is implemented as ASP's default negation \texttt{not}.

Drawing from Structural causal model, we impose the dependency graph between endogenous variables to be acyclic, so each $\varepsilon'$ (and then $\varepsilon$) also uniquely depends on $\mathcal{B}$ and $D$. Causal heads are unnegated event atoms, and the ground dependency graph includes both positive and negative event dependencies. Together with closed-world evaluation, this ensures a unique, consistent causal interpretation for each fixed $(\mathcal{B},D)$.

Relativizing the schema to the belief base $\mathcal{B}$ and decision $D$ makes this explicit:
\[
\varepsilon_{\mathcal{B},D} \leftarrow \theta \wedge \kappa_{\mathcal{B}} \wedge \alpha_{D} \wedge \varepsilon'_{\mathcal{B},D}
\]
Thus $\varepsilon_{\mathcal{B}_a,D}$ gives the expected consequences for agent $a$, while $\varepsilon_{\mathcal{B}_{env},D}$ represent the actual consequences. 
This is because, given a belief base and a decision, all consequences are uniquely determined. Although there may be uncertainty about which consequences occur, the causal theory itself is fixed common knowledge, so all uncertainty stems from the belief base.
Belief bases may disagree, but each fixed interpretation is two-valued.

\textit{Decision Theory} specifies which choices are available to the agent and when. By design there are \emph{choices} among actions, which opens multiple possible extensions, though constraints (incompatibilities between actions, forced reactions, situated opportunities) may apply.

A choice, written $\gamma$, is a triple:
\[
\gamma= \langle\mathcal{R},\min,\max\rangle
\]
where $\mathcal{R} \subseteq \mathcal{A}$ and $\min,\max\in\mathbb{N}$ with $\min\leq\max$. We write $\gamma.\mathcal{R}$, $\gamma.\min$, and $\gamma.\max$ for its three components.

The decision theory thus specifies how the choice are conditioned by a set of rules $\Pi$ of the form:
\[
\gamma \;\leftarrow\; \delta
\]
where $\delta= \theta \wedge \kappa \wedge \alpha \wedge \varepsilon$ is a decision condition, a conjunction of literals exactly as for causal rules. Thus, when $\delta$ holds, the choice $\gamma$ requires selecting between $\min$ and $\max$ actions from $\mathcal{R}$, the set of admissible actions.

Based on its beliefs and their expected consequences, the agent forms a \emph{decision} $D$, the set of actions it selects. Given a set of choices $\Gamma$, a decision $D$ respects $\Gamma$ if and only if:
\begin{enumerate}
\item $D\subseteq \mathop{\bigcup}\limits_{\gamma\in\Gamma} \gamma.\mathcal{R}$, and
\item $\forall \gamma \in \Gamma,\quad
\gamma.\min \;\le\; \bigl|\gamma.\mathcal{R} \cap D\bigr| \;\le\; \gamma.\max$
\end{enumerate}

As a result, given the decision theory $\Pi$ and a belief base $\mathcal{B}$, the admissible decisions for $a$ are those $D$ respecting the triggered choices
\[
\Gamma_{\mathcal{B},D}
= \bigl\{\,\gamma \;\bigm|\; (\gamma \leftarrow \delta)\in\Pi\ \text{and}\ \delta_{\mathcal{B},D} \text{~holds} \,\bigr\}.
\]

A natural question concerns the belief base $\mathcal{B}$ in $\Gamma_{\mathcal{B},D}$. As decisions are taken by the agent, it should be based on its own beliefs and their expected consequence. We thus use $\mathcal{B}=\mathcal{B}_{a}$ to evaluate which choices are actually done. 

At execution time, however, feasibility depends on the environment's state $\mathcal{B}_{\texttt{env}}$, via $\delta_{\mathcal{B}_{\texttt{env}},N}$, and a mismatch between plan and execution may trigger contradiction (for example if an unanticipated choice is required and cannot be ignored) and force the agent to revise its beliefs and replan its decision from that point onward. Such cases are easily avoided when no forced reaction (a choice with $\gamma.\min>0$) depends on uncertain knowledge, and we leave them for future work.

\subsubsection{Instantiation in Sartre's \textit{The Wall}.}

Here the decider is Pablo \texttt{p}. 
His belief base $\mathcal{B}_{\texttt{p}}$ and the true state of the environment $\mathcal{B}_{\texttt{env}}$ are set to:

\smallskip
\noindent
\resizebox{\columnwidth}{!}{%
{
\begin{tabular}{@{}l l@{}}
$\mathcal{B}_{\texttt{env}} =
\begin{cases}
  \texttt{at(r, cemetery)} \\
  \texttt{hostile(q,r)} \\
  \texttt{hostile(q,p)} \\
  \texttt{murderer(q)} \\
  \texttt{credible(at(r, family), q)} \\
  \texttt{credible(at(r, cemetery), q)}
\end{cases}$
&
$\mathcal{B}_{\texttt{p}_{0}} =
\begin{cases}
  \texttt{at(r, family)} \\
  \texttt{hostile(q,r)} \\
  \texttt{hostile(q,p)} \\
\texttt{murderer(q)} \\
  \texttt{credible(at(r, family), q)} \\
  \texttt{credible(at(r, cemetery), q)} \\
  \texttt{credible(at(r, madrid), q)}
\end{cases}$ \\
\end{tabular}
}}
\smallskip

Full instantiation of the causal theory can be found alongside our code, but we give here some examples of causal rules (omitting domain predicate from $\theta$ by considering $X$,$Y$,$Z$ are persons and $L$ locations):

\begin{itemize}\small
\item[$\bullet$] \texttt{check($X$,$Y$,$L$,$Z$)} $\leftarrow $ \texttt{credible(at($Z$,$L$}),$X$), \texttt{utter}($Y$,$X$,$N$, \texttt{inform(at}($Z$,$L$)))

\item[$\bullet$] \texttt{kill($X$,$Y$,$Z$)} $\leftarrow $ \texttt{at($Z$,$L$)}, \texttt{check($X$,$Y$,$L$,$Z$)}
\item[$\bullet$] \texttt{escaped($X$,$Y$)} $\leftarrow $ \texttt{evade($X$,$Y$)}, \texttt{check($Y$,$X$, madrid,$Z$)}
\item[$\bullet$] \texttt{harm($X$,$Y$,uncoveredLie)} $\leftarrow $ $\neg$\texttt{at($Z$,$L$)}, \texttt{check($X$,$Y$,$L$,$Z$)}, $\neg$\texttt{escaped($Y$,$X$)}
\end{itemize}

To instantiate the decision theory on Sartre's story, recall the Falangists' reference question from the Basic Language (subsection \ref{ssec:basiclanguage}), which asks for Ramon's location:
\[
\delta=\texttt{utter(q,p,}N\texttt{,query(at(r,}x\texttt{)))}
\]
As a decision condition, $\delta$ triggers a single choice $\gamma=\langle\mathcal{R},1,1\rangle$ whose action set $\mathcal{R}\subseteq\mathcal{A}$ gathers Pablo's admissible replies: the informings $\texttt{utter(p,q,}N\texttt{,inform(at(r,}L\texttt{)))}$ for each candidate location $L$, together with \texttt{silence(p,q)}. Since $\min=\max=1$, Pablo gives a single answer or remains silent, and any answer must fall within these admissible options. A further choice is situated rather than immediate: should the Falangists leave for Madrid to verify the claim, the resulting opportunity licenses \texttt{evade(p,q)}, by which Pablo may escape.

\subsection{Speech Acts Analysis}
\label{ssec:speechanalysis}


Next layer in our architecture is the speech act analysis. 
Classical approach of speech act theory, as defined in FIPA-ACL~\cite{fipa2002}, is normative: performative are characterized in the Semantic Language (SL) by a feasibility precondition (FP) and rational effect (RE)~\cite{cohenlevesque1990,pittmamdani1999} that are not physical constraints, but rather normative expectations between honest agents.
Typically, for the main assertive performative \texttt{inform}, which is our focus here as we study lying, we have:
\begingroup\footnotesize
\[
\text{FP}:\ B_{\texttt{p}}\sigma \wedge \neg B_{\texttt{p}}\!\left(\mathit{Bif}_{\texttt{q}}\sigma \vee \mathit{Uif}_{\texttt{q}}\sigma\right),
\qquad
\text{RE}:\ B_{\texttt{q}}\sigma,
\]
\endgroup
with $\mathit{Bif}_{\texttt{q}}\sigma \equiv B_{\texttt{q}}\sigma \vee B_{\texttt{q}}\neg\sigma$ and $\mathit{Uif}_{\texttt{q}}\sigma \equiv U_{\texttt{q}}\sigma \vee U_{\texttt{q}}\neg\sigma$, where $U$ is the uncertainty modality. The FP is a sincerity clause requiring $B_{\texttt{p}}\sigma$ with an additional operational requirement to avoid informing another of something it already knows, and the RE is the intended belief change $B_{\texttt{q}}\sigma$. In our model, external agent's beliefs are integrated to the previous layer, so  operational precondition can be simulated in the decision theory while the intended effect is manifested by the events it triggers through the causal model and can be weakened through the predicate \texttt{credible}$(\sigma,\texttt{q})$. 

In this layer, given a decision $D$, we evaluate a content $\sigma$ against a belief base: $\sigma_{\mathcal{B}_{\texttt{p}},D}$ means Pablo believes $\sigma$ and $\neg\sigma_{\mathcal{B}_{\texttt{p}},D}$ that he does not, while $\sigma_{\mathcal{B}_{\texttt{env}},D}$ means the environment holds $\sigma$ true. 
On this basis, we define four predicates capturing the speaker's honesty and the truth status of the utterance: \texttt{honest}, \texttt{dishonest}, \texttt{truth}, and \texttt{falsity}. A speech act is honest if the uttered content is believed by the speaker, which is exactly the sincerity clause of the \texttt{inform} feasibility precondition:
\[
\texttt{honest}(\texttt{p}, \sigma) \leftarrow \texttt{utter}(\texttt{p},\texttt{q},N,\texttt{inform}(\sigma)),\ \sigma_{\mathcal{B}_{\texttt{p}},D}
\]
It is dishonest if the speaker does not believe what they assert:
\[
\texttt{dishonest}(\texttt{p}, \sigma) \leftarrow \texttt{utter}(\texttt{p},\texttt{q},N,\texttt{inform}(\sigma)),\ \neg \sigma_{\mathcal{B}_{\texttt{p}},D}
\]

Under closed-world completion, content not supported by the speaker's beliefs is treated as false, not unknown. The model therefore cannot represent a speaker who believes neither $\sigma$ nor $\neg\sigma$, and so cannot express bullshit in the sense of~\cite{sakama2015formal}, namely uttering $\sigma$ in that epistemic state.

In our setting, truth and falsity are instead evaluated with respect to the environment's belief base:

\begin{center}
\begingroup
\setlength{\tabcolsep}{0pt}
\begin{tabular}{l}
$\texttt{truth}(\texttt{p}, \sigma) \leftarrow \texttt{utter}(\texttt{p},\texttt{q},N,\texttt{inform}(\sigma)),\ \sigma_{\mathcal{B}_{\texttt{env}},D}$\\
$\texttt{falsity}(\texttt{p}, \sigma) \leftarrow \texttt{utter}(\texttt{p},\texttt{q},N,\texttt{inform}(\sigma)),\ \neg \sigma_{\mathcal{B}_{\texttt{env}},D}$
\end{tabular}
\endgroup
\end{center}

These base predicates allow us to define four compound categories of speech acts that combine epistemic stance with truth-conditional evaluation:
\begin{center}
\begingroup
\setlength{\tabcolsep}{0pt}
\begin{ttfamily}
\begin{tabular}{l}
objective\_truth(\texttt{p},$\sigma$) $\leftarrow$ honest(\texttt{p},$\sigma$), truth(\texttt{p},$\sigma$) \\
erroneous\_truth(\texttt{p},$\sigma$) $\leftarrow$ honest(\texttt{p},$\sigma$), falsity(\texttt{p},$\sigma$) \\
objective\_lie(\texttt{p},$\sigma$) $\leftarrow$ dishonest(\texttt{p},$\sigma$), falsity(\texttt{p},$\sigma$) \\
erroneous\_lie(\texttt{p},$\sigma$) $\leftarrow$ dishonest(\texttt{p},$\sigma$), truth(\texttt{p},$\sigma$)
\end{tabular}
\end{ttfamily}%
\endgroup
\end{center}


\subsection{Ethical Evaluation}

To formally evaluate the morality of decisions involving speech-act utterances, we define rules that determine whether a decision $D$ is \textit{permissible} or \textit{impermissible} under a moral theory. Each theory encodes its own normative criteria, whether maxims, principles, or utility functions, and relies on distinct assumptions about which factors are morally relevant. 

To concisely represent possible states of the world (with all unfolding consequences) considered in moral evaluation, we define a set of \textit{situations} $\mathbb{S} = \{ \mathcal{S}, \mathcal{S}', \mathcal{S}'', \ldots \}$. 
Since the causal model is acyclic, consequence are deterministic and thus each possible decision $D$ is associated to a single situation $\mathcal{S}$, from which we can recover both the expected and real consequences. 

We say that an ethical evaluation is \emph{local} if a decision can be fully assessed by considering only its associated situation (the choices actually made), and \emph{global} if it requires comparing that situation
with others (what else the agent could have done). Computationally, local evaluation can thus be carried out as decisions are generated and their consequences computed, whereas global criteria require an additional layer explicitly representing the situations to be compared.

Additional meta-variables sets are the ethical specification related to each approach and will be described in their respective parts. We now detail the different theories and their criterion.

\textit{Consequentialism} deems a speech-act utterance, like any action, permissible if it leads to the best possible outcomes and impermissible otherwise. The two variants differ in evaluative perspective. \texttt{\small consequentialism1} is expected-utilitarian, assessing anticipated utility from the agent's beliefs, with $\mathcal{B}_{\texttt{p}}$ as epistemic reference. \texttt{\small consequentialism2} is real-utilitarian, assessing actual consequences through the environment's idealized belief base $\mathcal{B}_{\texttt{env}}$.

Central to utilitarianism is the \emph{value theory} \cite{sep-value-theory} which defines a utility function mapping each triggered event to an integer:
\[
\text{utility} : \mathcal{E} \to \mathbb{Z}
\]
where $\text{utility}(\varepsilon)$ measures the moral value (benefit or
harm) of each possible consequence $\varepsilon \in \mathcal{E}$. The most classical value theory is \emph{hedonism}, which assesses a state by the
well-being or pleasure of all involved parties: truths and lies then
matter only through their effect on overall pleasure. In more pluralistic views such as G.E.~Moore's~\cite{Moore1903-MOOPE}, for which knowledge is
an intrinsic value, a lie carries a disvalue of its own through its
epistemic consequences, which would then have to be explicitly represented
in $\mathcal{E}$. In our running example, as practical consequences are life-or-death, we consider that such badness will be negligible compared to the utility given to \texttt{kill($\cdot,\cdot,\cdot)$} (-4) and \texttt{harm($\cdot,\cdot,\cdot)$} (-3).

For a situation $\mathcal{S}$ evaluated under a belief base $\mathcal{B}$, the total utility can be obtained locally as the sum of the utilities of the resulting consequences:
\[
\text{TotalUtility}(\mathcal{S}, \mathcal{B}) = \sum_{\varepsilon \in \mathcal{E}_{\mathcal{S},\mathcal{B}}} \text{utility}(\varepsilon)
\]
where $\mathcal{E}_{\mathcal{S},\mathcal{B}}$ is the set of consequences arising in situation $\mathcal{S}$ under belief base $\mathcal{B}$.

Let $\mathbb{S}$ be the set of candidate situations. The morally optimal plan maximizes total utility. As this requires comparing each situation's total utility, it is a global theory.

In \texttt{\small consequentialism1}, expected utility is computed from the agent's subjective belief base $\mathcal{B}_{\texttt{p}}$:
\[
\mathcal{S} \text{ is permissible \textit{iff} } \mathcal{S}\in\arg\max_{\mathcal{S}' \in \mathbb{S}}\text{TotalUtility}(\mathcal{S}', \mathcal{B}_{\texttt{p}})
\]
so $\text{TotalUtility}(\mathcal{S}', \mathcal{B}_{\texttt{p}})$ is the utility the agent believes would follow from the actions chosen in $\mathcal{S}'$. This is forward-looking, internalist evaluation.

In \texttt{\small consequentialism2}, actual utility is computed from the environment's idealized belief base $\mathcal{B}_{\texttt{env}}$, representing the objective consequences:
\[
\mathcal{S} \text{ is permissible \textit{iff} } \mathcal{S}\in\arg\max_{\mathcal{S}' \in \mathbb{S}} \text{TotalUtility}(\mathcal{S}', \mathcal{B}_{\texttt{env}})
\]
so $\text{TotalUtility}(\mathcal{S}', \mathcal{B}_{\texttt{env}})$ returns the actual utility produced in $\mathcal{S}'$. This is retrospective, externalist evaluation.

\textit{Deontologism} evaluates the permissibility of decisions by whether they comply with absolute moral maxims, as in Kant's categorical imperative. 
We consider a set $\mathbb{M}$ of deontological maxims: $\mathbb{M}=\{ \texttt{maxim1}, \texttt{maxim2}\}$, relevant to the problem. 
Here, \texttt{maxim2} is the absolute `do not lie' maxim defended by Kant in the controversy. We also add a `do not enable killing' maxim (\texttt{maxim1}) to show the tension between the different imperatives in this dilemma. 
For each maxim, violations are defined by rules of the form:
\[
\mathrm{violated}_{\mathcal{S}}(\texttt{maxim}) \leftarrow \delta
\]
where $\texttt{maxim}\in\mathbb{M}$, and $\delta$ specifies when the maxim is violated. In our setting:

\begin{center}
\begingroup
\begin{tabular}{l}
$\mathrm{violated}_{\mathcal{S}}(\texttt{maxim1}) \leftarrow \texttt{kill($\cdot$,p,$\cdot$)}$\\
$\mathrm{violated}_{\mathcal{S}}(\texttt{maxim2}) \leftarrow \texttt{dishonest(p,$\cdot$)}$
\end{tabular}
\endgroup
\end{center}

Deontologism admits no exceptions, as maxims must be upheld unconditionally. An action violating a maxim is thus impermissible, otherwise permissible.

\textit{Principialism} evaluates actions by their compliance with normative
principles, gathered in $\mathbb{P}=\{\texttt{principle0},
\texttt{principle1}, \texttt{principle2}\}$; unlike deontologism, it admits
exceptions under contextual conditions. Both variants combine the honesty
principle that distinguishes them (\texttt{principle1} or
\texttt{principle2}) with the shared non-maleficence \texttt{principle0}, which mirrors \texttt{maxim1}: it is violated when the
agent anticipates that the utterance enables a \texttt{kill} event. It is \texttt{principle0} that renders the honest options impermissible in Table~\ref{tab:eval}, where \texttt{objective\_truth} (counterfactual for
Pablo, who believes \texttt{at(r,family)}) is included for completeness. We now formalize \texttt{\small principialism1} and \texttt{\small principialism2}.

In both variants, a principle is violated when a condition $\delta$ holds:
\[
\mathrm{violated}_{\mathcal{S}}(\texttt{principle}) \leftarrow \delta
\]
where $\texttt{principle}\in\mathbb{P}$, and $\delta$ is a complete condition. Recall that the two variants differ in the exception applied to \texttt{principle1} and \texttt{principle2}, respectively. \texttt{\small principialism1} excuses a violation of \texttt{principle1} under a local condition \(\delta'\):
\[
\mathrm{exception}_{\mathcal{S}}(\texttt{principle1}) \leftarrow \delta'
\]
\texttt{\small principialism2} instead requires a global evaluation, in which the exception to \texttt{principle2} compares outcomes across counterfactual situations. Here $\delta'$ is a local, situation-specific condition, whereas its global counterpart $\Delta'$ compares multiple situations:
\[
\mathrm{exception}_{\mathcal{S}}(\texttt{principle2}) \leftarrow \Delta'
\]
To make this concrete, consider Pablo (\texttt{p}) in Sartre's \textit{The Wall}, and write $\alpha(\texttt{L})$ for the informing \texttt{utter(p,q,}\textit{N}\texttt{,inform(at(r,L)))}. He believes Ramon (\texttt{r}) is at his family's house, and must choose between the dishonest $\alpha(\texttt{L'})$, placing \texttt{r} at a location \(\texttt{L'}\) (Madrid or the cemetery), and the honest $\alpha(\texttt{L})$, placing \texttt{r} at \(\texttt{L}\) (the family's house), which Pablo believes correct.

Under \texttt{\small principialism2}, the lie is permissible if being honest, that is, uttering \(\alpha(\texttt{L})\), would lead to worse consequences. This counterfactual comparison is:
\[
\begin{aligned}
\mathrm{exception}_{\mathcal{S}'}( \texttt{principle2}) \leftarrow{}&\ \alpha(\texttt{L})_{\mathcal{S}},\ \alpha(\texttt{L'})_{\mathcal{S}'},\ \texttt{L} \neq \texttt{L'},\ \texttt{at}(\texttt{r}, \texttt{L})_{\mathcal{B}_{\texttt{p}}},\\
&\ \mathrm{TotalUtility}(\mathcal{S}', \mathcal{B}_{\texttt{p}}) > \mathrm{TotalUtility}(\mathcal{S}, \mathcal{B}_{\texttt{p}})
\end{aligned}
\]
The dishonest utterance \(\alpha(\texttt{L'})\) in situation \(\mathcal{S}'\) is excused if, in the alternative situation \(\mathcal{S}\), being honest (uttering \(\alpha(\texttt{L})\), the location Pablo believes correct) yields lower total utility. The exception borrows a utilitarian criterion to judge whether the counterfactual situation is worse. This supports the permissibility of both of Pablo's lies under \texttt{\small principialism2} in Table~\ref{tab:eval} (see subsection~\ref{ssec:moraltheo}).

Finally, impermissibility is defined: an action is impermissible if it violates a principle and no exception applies,
\[
\mathrm{impermissible}_{\mathcal{S}}(\alpha) \leftarrow \mathrm{violated}_{\mathcal{S}}(\alpha,\texttt{principle}),\ \mathrm{not}\ \mathrm{exception}_{\mathcal{S}}(\alpha,\texttt{principle})
\]
and permissible whenever it is not impermissible,
\[
\mathrm{permissible}_{\mathcal{S}}(\alpha) \leftarrow \mathrm{not}\ \mathrm{impermissible}_{\mathcal{S}}(\alpha)
\]
so that in \texttt{\small principialism} permissibility holds by default, unless a principle is violated with no valid exception.

These rules encode defeasible norms through explicit exceptions rather than a separate priority relation. The architecture also specifies their causal and epistemic conditions and compares alternative decisions.


\section{ASP Evaluation}
\label{sec:evalframework}

We realise the logical framework in ASP, using Sartre's \textit{The Wall} as an illustration. The implementation uses Python 3.8+, Clingo\footnote{\url{https://potassco.org}} as solver, and Clyngor\footnote{\url{https://pypi.org/project/clyngor/0.0.13/}} for the ASP-Python interface. The codebase can be accessed at: \url{https://github.com/benjaminicard/prima2026}

The program separates the classification and evaluation rules from the scenario descriptions, given as belief-base facts and a utility assignment: only these
facts change across the base scenario and its alternatives, the rules being reapplied verbatim, a split checkable in the source. Each answer set is one
admissible decision sequence, pairing a reply among the informings and silence
with the situated opportunity to evade; utilitarian optima aggregate utilities per candidate situation and compare across them, rather than using weak-constraint optimization.

The evaluation on Sartre's \textit{The Wall} validates the logical formalization on two counts. Firstly, it reproduces the intended epistemic classifications of speech acts, with Pablo's ``Madrid'' reply derived as an \texttt{objective\_lie} and his ``cemetery'' reply as an \texttt{erroneous\_lie}. Secondly, these classifications feed the moral evaluation rules of moral theories, whose verdicts are consistent with established ethical analyses of lying~\cite{bok1978lying,carson2010lying}:

\begin{itemize}
\item[—] For \textbf{deontologism}, both Pablo's lies are categorically \textit{impermissible}, as they violate \texttt{maxim2} as a result of Pablo's dishonesty.

\item[—] For \textbf{principialism}, both lies are \textit{permissible}, but for distinct reasons. Under \texttt{\small{principialism1}}, both lies satisfy the exception that permits lying when the truth is not deserved. Under \texttt{\small{principialism2}}, both exceptions are granted by counterfactual reasoning, since telling the truth would have produced worse outcomes.

\item[—] For \textbf{consequentialism}, expected utility under \texttt{\small{cons\-equentialism1}} singles out ``Madrid'' together with the evasion it makes possible, the sole decision Pablo believes harmless. Under \texttt{\small{consequentialism2}}, evaluating actual utility, the ``Madrid'' lie is permissible but the ``cemetery'' one is not, since it ultimately leads to Ramon's death.
\end{itemize}

Where Table~\ref{tab:eval} evaluates each utterance on its own, the program compares complete decisions, silence and evasion included, so the consequentialist verdicts are taken over a larger set of situations and refine accordingly. On this base case, both principialist variants converge on permitting both lies, though for different reasons.

\section{Contrasting Variants}
\label{sec:discussion}

Two variants of the base scenario, each obtained by a single parameter change, locate where the agreement between theories breaks.

In \textbf{Alternative~1}, Pablo lies to the hostile Falangists that Ramon
is in Madrid. Unlike the original story, this helps both, as the Falangists
search distant Madrid, giving Pablo time to escape. Under
\texttt{\small principialism}, ``Madrid'' is as \textit{permissible} as
``Cemetery'' (\texttt{\small principialism1} by the Falangists' murderous intent,
\texttt{\small principialism2} because honesty would be worse than either
lie). The equivalence breaks under \texttt{\small consequentialism1}: by
expected utility ``Madrid'' uniquely secures both survivals, whereas
``Cemetery'' still endangers Pablo, so it singles out ``Madrid'' as optimal.
Formally, Alternative~1 is the base instantiation plus a single belief-base
fact, \texttt{credible(at(r,madrid),q)} added to
$\mathcal{B}_{\texttt{env}}$, with zero rule changes.

In \textbf{Alternative~2}, Pablo faces not the Falangists but a good friend
who, though not a murderer by nature, has a private motive to kill Ramon.
Pablo lies that ``Ramon is at the cemetery.'' Here the variants diverge:
\texttt{\small principialism1} deems the lie impermissible, as a non-hostile
friend does not satisfy \texttt{principle1}'s exception, whereas
\texttt{\small principialism2} deems it permissible, since honesty would
cause greater harm, satisfying \texttt{principle2}. Formally,
Alternative~2 is the base instantiation minus the \texttt{murderer(q)}
facts (the interlocutor no longer forfeits the right to the truth), the
kill-enabling disposition being retained, again with zero rule changes.

This \textit{post hoc} analysis shows that theories converge on simple
cases, such as Pablo's two lies in the original story, but diverge once
contextual parameters vary: here, expected utility (Alternative~1) or the
interlocutor's status (Alternative~2), where the divergence marks the boundary between Constant's criterion, a right to the truth that murderous intent forfeits, and a consequence-based one, from which his position is distinct
(see Section~\ref{ssec:moralspeech}). Deontologism, by contrast, is unmoved
by both parameters, since \texttt{maxim2} is violated by the lie itself. Our framework does not settle the controversy but shows which of its criteria the parameters can move.


\section{Conclusion}
\label{sec:conclusion}

This paper presents a logic-based framework that treats speech acts, and not only physical actions, as first-class objects of moral evaluation: a belief-based agent architecture and a causal model derive the outcomes of each utterance, honest or deceptive, whose permissibility is then assessed under deontologism, consequentialism, and principialism. Separating rules from scenario parameters, it applies to any case whose
consequences can be foreseen by a causal model, in the line of Structural Equation Models~\cite{pearl2009causality} and prior ASP accounts of moral action~\cite{berreby2015modelling,bourgne2021ace}. Also realised in ASP, our framework is illustrated on Sartre's \textit{The Wall} and on two variants obtained by
parameter change alone, which formalize the Kant-Constant classical controversy and locate where its parties divide.

As immediate future work, we aim, first, to handle plan-execution
mismatches through observation-based action revision under weak-constraint optimization and, second, to cover further
performatives, notably
\textit{commissives}~\cite{searle1976,searlevanderveken1985} like
\texttt{commit} and \texttt{promise}, which the \texttt{utter} schema readily accommodates. In the longer term, finer-grained epistemic extensions, such as higher-order beliefs, deception as intended belief change, or common knowledge, would call for a genuinely modal treatment like dynamic epistemic logics of lying~\cite{vanditmarsch2014dynamics}.

\begin{credits}
\subsubsection{\ackname}
We thank two anonymous reviewers for their helpful comments and suggestions. BI, GB and JGG acknowledge the programs TRUSTEDNEWS (ANR-25-ASM2-0003) and THEMIS (grant agreements n°DOS022279400 and n°DOS022279500). We also thank the audience of the GDR RADIA 2025. 

\subsubsection{\discintname}
The authors have no competing interests to declare that are relevant to the content of this article.

\subsubsection{AI Declaration.}
Generative AI was used to improve the English wording and readability of the manuscript. The authors reviewed and revised the suggestions and take full responsibility for the final content.

\end{credits}

\newpage
\bibliographystyle{splncs04}
\bibliography{sample}


\end{document}